\documentclass[letterpaper, 10 pt, conference]{ieeeconf}  
\usepackage{graphicx}
\usepackage{xcolor}

\usepackage{amsmath}
\IEEEoverridecommandlockouts                              

\title{\LARGE \bf
Terrain characterization and locomotion adaptation in a small-scale lizard-inspired robot
}

\author{Duncan Andrews$^{1}$, Landon Zimmerman$^{1}$, Evan Martin$^{1}$, Joe DiGennaro$^{1}$, Baxi Chong$^{1}$\thanks{$^{1}$Duncan Andrews, Landon Zimmerman, Evan Martin, Joe DiGennaro, and Baxi Chong are with the Locomotion in Biology and Robotics (LiBR) Lab at Penn State University, University Park, PA 16802 {\tt\small \{dja5722, lcz5044, erm5808, jrd6327 baxi.chong\}@psu.edu, }}}

\begin{document}

\maketitle
\thispagestyle{empty}
\pagestyle{empty}

\begin{abstract}

Unlike their large-scale counterparts, small-scale robots are largely confined to laboratory environments and are rarely deployed in real-world settings. As robot size decreases, robot–terrain interactions fundamentally change; however, there remains a lack of systematic understanding of what sensory information small-scale robots should acquire and how they should respond when traversing complex natural terrains. To address these challenges, we develop a Small-scale, Intelligent, Lizard-inspired, Adaptive Robot (SILA Bot) capable of adapting to diverse substrates. We use granular media of varying depths as a controlled yet representative terrain paradigm. We show that the optimal body movement pattern (ranging from standing-wave bending that assists limb retraction on flat ground to traveling-wave undulation that generates thrust in deep granular media) can be parameterized and approximated as a linear function of granular depth. Furthermore, proprioceptive signals, such as joint torque, provide sufficient information to estimate granular depth via a K-Nearest Neighbors classifier, achieving 95\% accuracy. Leveraging these relationships, we design a simple linear feedback controller that modulates body phase and substantially improves locomotion performance on terrains with unknown depth. Together, these results establish a principled framework for perception and control in small-scale locomotion and enable effective terrain-adaptive locomotion while maintaining low computational complexity.
\end{abstract}

\section{INTRODUCTION}

The last decade has seen remarkable success in large-scale legged robotics, which employ well-coordinated sensor suites and control policies  and have demonstrated robust mobility across rugged terrains \cite{bostondynamics2024SpotSpecs, margolis2024RapidRL, lee2020LearningQuad, misenti2025ExperimentalEvaluation, fernandes2025grapevine_pruning_3d}. However, their height (often on the order of tens of centimeters, up to $\sim$70\,cm) \cite{bostondynamics2024SpotSpecs} limits their ability to access confined spaces, such as low-hanging vegetation in agriculture, collapsed structures, mineral deposits, or inside segments of pipes. Thus, there is a need for small-scale (on the order of 5 to 15\,cm tall) robots capable of navigating the environments where large-scale robots cannot reach. 

However, simply shrinking existing legged robots is not always a viable strategy because performance is not maintained when reducing size. At small scales, physical constraints on sensors and hardware cause issues for perception and control. For example, smaller onboard sensors typically have lower sensor resolution and are more susceptible to noise  \cite{mohammed2018MonolithicAccelerometers, kim2008OptimizationPixel}, which makes navigation challenging. Additionally, smaller motors produce less torque and are constrained by other factors like power consumption. Instead of generalizing/transferring strategies from larger robots, reliable navigation and locomotion at smaller scales demands new perception and control principles that account for the different constraints.

\begin{figure}[t]
    \centering
    \includegraphics[width=\linewidth]{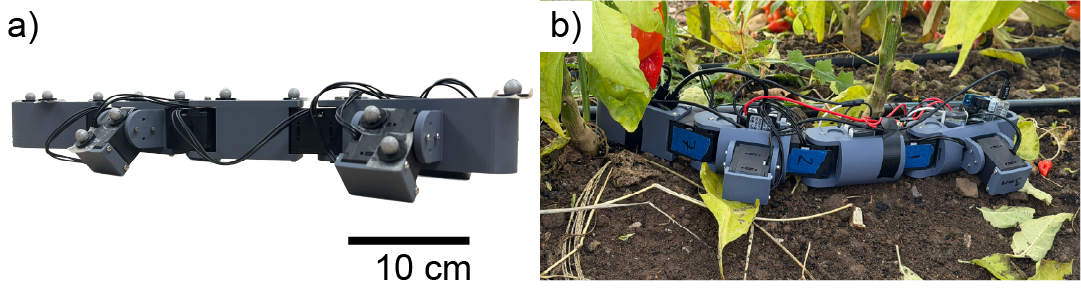}
    \vspace*{-5mm}
    \caption{\textbf{Small-scale, Intelligent, Lizard-inspired, Adaptive Robot (SILA Bot)} a) Lizard robot with 7 actuated servos: 3 body servos and 4 leg servos. b) Wirelessly configured lizard robot walking through soil with unknown depth at a farm.}
    \label{fig:Robot}
\end{figure}%

In addition to hardware constraints, small-scale robots face a different set of advantages and challenges when compared to larger platforms. Their low center of mass and belly contact with the ground provide inherent stability, making them less prone to tipping, which reduces the need for complex perception and stabilization techniques found in large-scale robots. Instead, the defining challenge at small scales is that terrain clutter (e.g., sticks, leaves, rubble, rocks, sand) is comparable to the size of the robot. As a result, locomotion is not limited by balance, but the ability to generate reliable self-propulsion over all kinds of clutter and through various complex terrains.

The propulsion mechanisms required for small-scale locomotion depend strongly on substrate physics. On flat, rigid ground, interactions are dominated by Coulomb friction, and forward motion primarily arises from coordinated limb retraction. Body movements assist locomotion by modulating load distribution and enhancing traction \cite{li2013TerradynamicsLegged}. In contrast, on flowable substrates, such as sand, the body itself can serve as a primary propulsor (similar to snakes \cite{astley2015Sidewinding} or snake-inspired robots \cite{hatton2013GeometricSelfProp}). Undulatory motion generates thrust through resistive forces within the granular media, even when limb traction is reduced. Given the  distinct thrust generation mechanisms on different terrains, two capabilities are essential for small-scale robots to move reliably in real-world environments: (1) environmental awareness to estimate terrain properties such as depth, geometry, or clutter distribution, and (2) appropriate coordination of belly–terrain interaction to modulate thrust generation.

For (1), vision is a common solution for terrain perception in robotics. However, it is inherently limited in detecting subsurface properties such as granular depth or yield characteristics, which are not directly visible \cite{jiang2025SafeActiveNavigation, brooks2007SelfSupervised}. Moreover, for low-profile robots operating close to the ground, cameras suffer from a restricted field of view and frequent occlusion by terrain features. For (2), there has been a lack of systematic frameworks to determine how body–terrain interaction should be modulated across different substrates. As a result, small-scale robots often rely on heuristics rather than substrate-aware propulsion strategies \cite{li2013TerradynamicsLegged, jiang2025SafeActiveNavigation}.

To address these challenges, we draw inspiration from biology. Lizards are among the most successful vertebrates and exhibit remarkable diversity in locomotion strategies across terrains \cite{uetz2020ReptileDatabase, pianka2003lizards}, ranging from bipedal running and quadrupedal trotting on solid ground to body-driven undulation on flowable substrates \cite{chong2022CoordinatingLizard, bergmann2017SandBetweenToes}. Inspired by this adaptability, we develop a Small-scale, Intelligent, Lizard-inspired, Adaptive Robot (SILA Bot) (Fig.~\ref{fig:Robot}) and establish a proprioceptive perception–control framework for locomotion across granular media (GM) of varying depths. Specifically, we show the optimal body coordination transitions from standing-wave to traveling-wave and is well-approximated by a linear function of depth over 0–40\,mm. Despite substantial environmental and hardware noise, onboard load measurements alone enable terrain depth classification ranging from 0 to 40\,mm with 95\% accuracy. Leveraging this relationship, we design a simple linear feedback controller that modulates body phase offset, significantly improving locomotor performance (up to 40\% when compared to a feedforward baseline) on terrains with unknown granular depth. Together, these results establish a principled framework for substrate-aware propulsion in small-scale robots and provide design guidelines for future terrain-adaptive systems operating in complex environments.



\section{EXPERIMENT SET-UP}

\subsection{Robot Design}

The SILA Bot uses Dynamixel XL-430-W250-T servo motors (Robotis), which are connected by 3D printed four PLA body segments. The segments are designed to allow a large range ($\pi/2$ radians for the body bending joints and $\pi$ radians for the leg shoulder elevation joints) of motion of all servo motors. SILA Bot stands approximately 5\,cm tall, and 45\,cm long. Robotis offers a Software Development Kit (SDK) that allows for writing and reading data to and from each servo motor. We use the SDK with MATLAB (MathWorks) to send a single command to all servos at the same time to minimize unwanted delays. 

\subsection{Conducting Experiments}
\begin{figure}
    \centering
    \includegraphics[width=\linewidth]{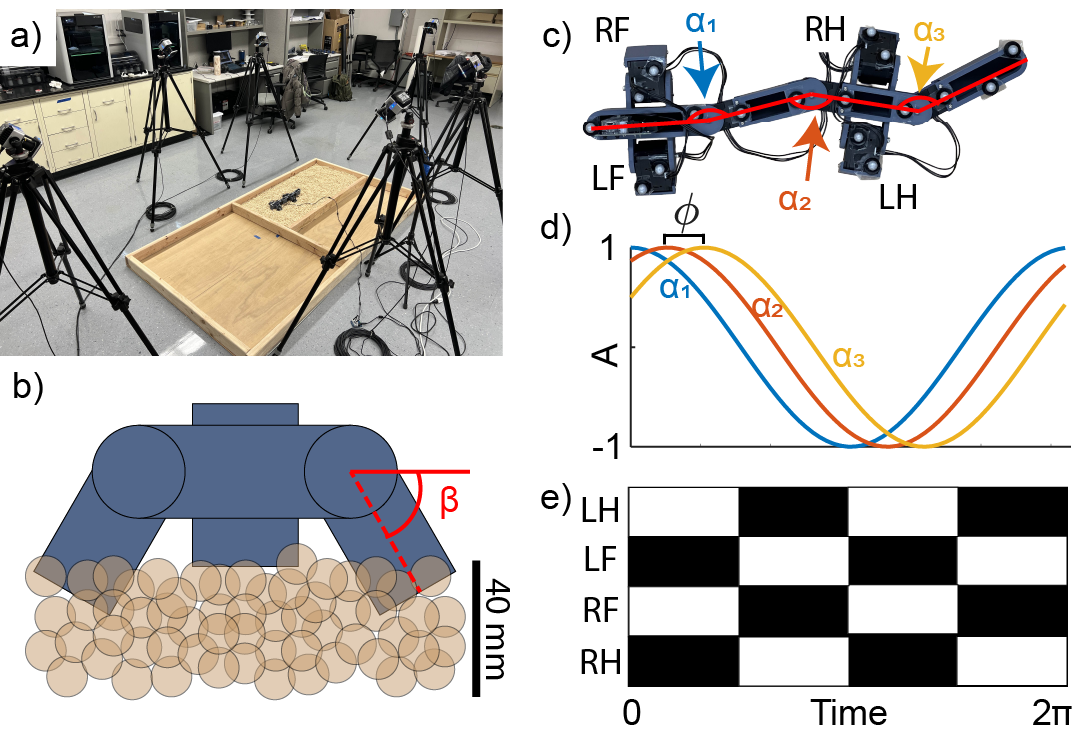}
    \vspace*{-5 mm}
    \caption{\textbf{Experimental setup and gait prescription.} 
    (a) Robot test arena consisting of an adjustable area containing granular media, surrounded by Vicon Vero motion-capture cameras for kinematic tracking. 
    (b) Schematic of the robot traversing granular media. 
    (c) Body joint angles $\alpha_1$–$\alpha_3$ defined along the body of the SILA Bot. L, R, F, and H stand for left, right, fore, and hind legs of the robot.
    (d) Cosine functions used to prescribe body joint trajectories; $A$ denotes the amplitude, and $\phi$ denotes the phase offset between consecutive joints. 
    (e) Limb contact sequence for LH, LF, RF, and RH legs of the robot. Shaded regions indicate ground contact. The robot employs a 50\% duty-cycle trotting gait in the current configuration.}
    \label{fig:Experiment}
\end{figure}

The lab space is outfitted with six Vicon Vero motion tracking cameras, capturing data at 100\,Hz, which we position around a test area shown in Fig.~\ref{fig:Experiment}.a. The robot is fitted with at least two reflective spheres on each rigid body that can be tracked by the cameras and recorded with sub-millimeter accuracy. To develop a controller for complex terrain, we test and record the robot on both flat ground and in a section of 10\,mm diameter wooden beads of varying depths, up to 40\,mm, illustrated in Fig.~\ref{fig:Experiment}.b. For this paper, we test at 0\,mm (flat ground), 20\,mm, and 40\,mm and the transition from flat ground to 40\,mm.  

\subsection{Gait Prescription}
We denote the three joint angles along the body as $\alpha_1$–$\alpha_3$, shown in Fig.~\ref{fig:Experiment}.c. We use sinusoidal functions to prescribe body joint angle movements and define body phase offset, $\phi$, as the phase difference between consecutive body joints (Fig.~\ref{fig:Experiment}.d). Using $\phi$, we can define the movement of the body motors of SILA Bot by:
\begin{equation}
    \alpha_n(t) = A\cos{(\omega t+(n-1)\phi)}
\end{equation}
\noindent where $A$ is the amplitude (fixed to 1 radian unless otherwise noted), $\omega$ is frequency (fixed to 1 radian/s unless otherwise noted) $t$ is the time step from 0 to $2\pi$ , $n$ is joint number, and $\phi$ is the body phase offset. A negative body phase offset induces a traveling wave in the robot, propagating from head to tail. A phase offset of 0 (meaning all motors move the same amount at each time step) corresponds to a standing wave \cite{chong2022CoordinatingLizard}.

We command a simple trotting gait~\cite{hildebrand1977AsymmetricalGaits} for the leg contact patterns (Fig.~\ref{fig:Experiment}.e) by prescribing the leg shoulder joints according to:
\begin{equation}
    \beta_i(t)  = \begin{cases}
    \beta_{land} & \text{if} \  \text{stance} \\
    \beta_{lift} & \text{otherwise}
\end{cases}
\end{equation}
\noindent for a leg in position $i$, where $\beta$ is the angle below horizontal (Fig.~\ref{fig:Experiment}.b). $\beta_{land}$ is set to $\pi/3$ radians, and $\beta_{lift}$ is set to 0 radians, unless otherwise noted. When SILA Bot is in the stance phase, the leg is lowered and is in contact with the ground. Otherwise, the leg is commanded to be lifted.

\subsection{Data Collection}
During data collection, Vicon cameras record $x$, $y$, and $z$ positions (the $x$ axis is defined as the long length of the test arena and the $y$ axis along the short) for each marker across all frames of the recording. We process this data using MATLAB and calculate the robot's body position as the average of all the points for a given frame. Dynamixel motors have built in functionality to read load as a percentage based on the linear relationship: $\tau=K_t*I$, Where $K_t$ is a constant and $I$ denotes current drawn by the servo motors. We use the load as an analog for torque. Each servo motor's load is recorded as a timeseries. However, because the load value is inferred from current and based on Pulse Width Modulation signals, it is prone to noise (with a measured coefficient of variation of$\sim13\%$). To address the fluctuations in the raw value, we apply a first-order low-pass filter (exponential smoothing) \cite{oppenheim2010DiscreteSignal} to the load signal.


\section{RESULTS}

\subsection{Effective Gaits on Different Terrains}
\begin{figure}
    \centering
    \includegraphics[width=\linewidth]{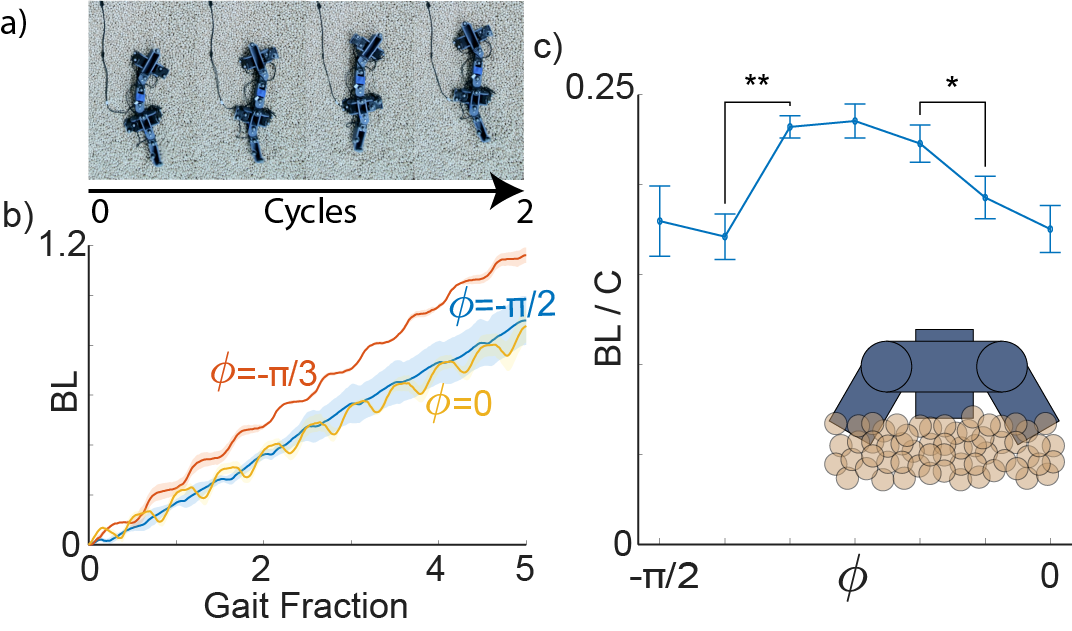}
    \vspace*{-5mm}
    \caption{\textbf{Deep granular media: body phase offset and locomotion speed.} 
    (a) Snapshots of SILA Bot executing a gait with body phase offset $\phi = -\pi/3$ on deep granular media. 
    (b) Experimentally tracked displacement as a function of gait fraction over five cycles. Three gaits are compared (yellow: $\phi = 0$; red: $\phi = -\pi/3$; blue: $\phi = -\pi/2$). 
    (c) Average forward speed (units: Body length per cycle, BL/C) as a function of body phase offset. Statistically significant differences (Paired T-Test) are observed between $\phi = -5\pi/12$ and $\phi = -\pi/3$ ($p < 0.01$), and between $\phi = -\pi/3$ and $\phi = -\pi/12$ ($p < 0.05$). The inset shows the robot traversing 40\,mm-deep granular media.}    
    \label{fig:40Speed}
\end{figure}

By testing SILA Bot on a variety of terrains, we observe that body phase offset correlates significantly with speed. We define speed by the number of body lengths that the robot travels per cycle (BL/C). First, we look at performance in deep GM. Fig.~\ref{fig:40Speed}.a depicts SILA Bot locomoting in 40\,mm deep wooden beads. We compared the recorded displacement as a function of gait fraction across three trials for three gaits with body phase offsets of $\phi = 0$, $\phi = -\pi/3$, and $\phi = -\pi/2$ for five cycles (Fig.~\ref{fig:40Speed}.b). In Fig.~\ref{fig:40Speed}.c, we plot the speed as a function of body phase offset, and we notice that $\phi=-\pi/3$ is significantly faster. 

\begin{table}[b]
    \caption{Best-performing Body Phase Offsets} 
    \centering 
    \begin{tabular}{|c|c|c|}
        \hline 
        0\,mm Depth & 20\,mm Depth & 40\,mm Depth \\ 
        \hline 
        $\phi^*=0$ & $\phi^*=-\pi/6$ & $\phi^*=-\pi/3$  \\ 
        \hline

    \end{tabular}
    \label{table:BestPhi} 
\end{table}

\begin{figure}
    \centering
    \includegraphics[width=\linewidth]{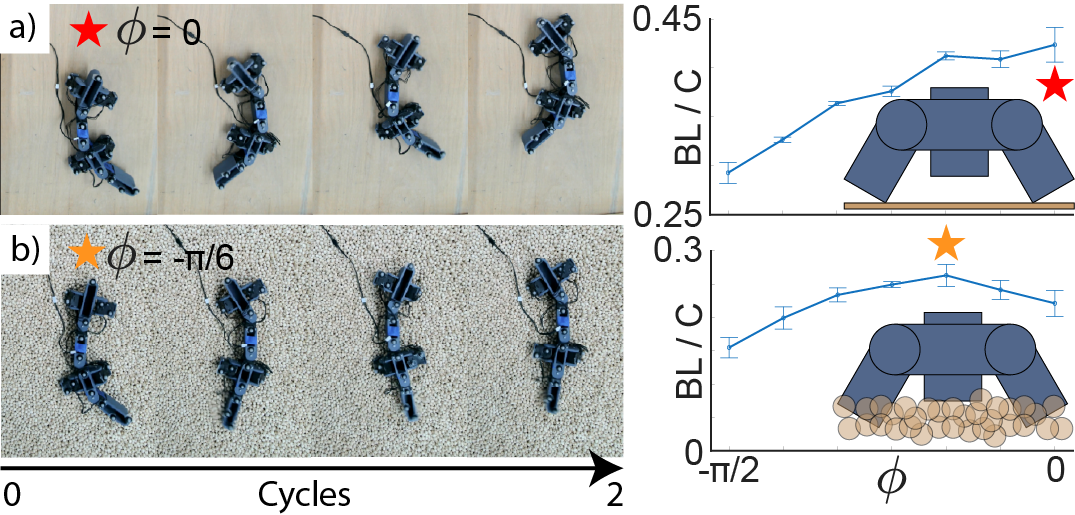}
    \vspace*{-5mm}
    \caption{\textbf{Shallow granular media: body phase offset and locomotion speed.} 
    (a) (\textit{left}) Snapshots of the SILA Bot executing a gait with body phase offset $\phi = 0$ on flat ground. 
    (\textit{right}) Average forward speed (units: Body length per cycle BL/C) of the SILA Bot on flat ground. 
    Inset: illustration of the robot traversing flat terrain. The red star corresponds to the best body phase, depicted \textit{left}.
    (b) (\textit{left}) Snapshots of the SILA Bot executing a gait with body phase offset $\phi = -\pi/6$ on shallow granular media (depth = 20\,mm). 
    (\textit{right}) Average forward speed (units: BL/C) on shallow granular media. 
    Inset: illustration of the robot traversing 20\,mm-deep granular media. The orange star corresponds to the best body phase, depicted \textit{left}.}     
    \label{fig:BestPhi}
\end{figure}

Next, SILA Bot is tested on flat ground shown in Fig.~\ref{fig:BestPhi}.a and 20\,mm deep wooden beads in Fig.~\ref{fig:BestPhi}.b. The photos of SILA Bot depict its best performance ($\phi=0$ and $\phi=-\pi/6$ for bead depths of 0 and 20\,mm, respectively). From this, a trend emerges: on flat ground, $\phi=0$ performs the best. In 20\,mm deep wooden beads $\phi=-\pi/6$ performs best, and at 40\,mm, $\phi=-\pi/3$ performs best. The best body phase offset for each depth is recorded in Table~\ref{table:BestPhi}. From Table~\ref{table:BestPhi}, we posit that for the bead depth in the range between 0 and 40mm, the optimal body phase offset $\phi^*$ for SILA Bot can be approximated by 
\begin{equation}
    \phi^* = -\frac{\pi}{120}d,
\end{equation}
\noindent where $d$ is the bead depth in millimeters. 

\subsection{Resistive Force and Coulomb Friction Models}

To better understand how proprioceptive sensing can be leveraged for terrain characterization, we develop a simplified physical model of robot–terrain interaction. The exact ground reaction forces in shallow granular media (GM) remain insufficiently characterized in the literature. In contrast, the two limiting cases - resistive force theory (RFT) for deep, fully yielded GM and dry Coulomb friction for rigid flat ground - have been extensively studied \cite{li2013TerradynamicsLegged}. We therefore posit that reaction forces in shallow GM are bounded by these two extremes: (i) Coulomb friction on rigid ground and (ii) granular RFT applicable to deep substrates. To simplify our analysis, we approximate shallow granular interaction as a linear combination of these limiting behaviors. While this assumption substantially simplifies the underlying mechanics, our objective is not precise force prediction but rather to extract qualitative insights into which proprioceptive signals are informative for terrain differentiation. Quantitative terrain classification is subsequently achieved through experiments and data-driven methods, as described later.

\begin{figure}
    \centering
    \includegraphics[width=\linewidth]{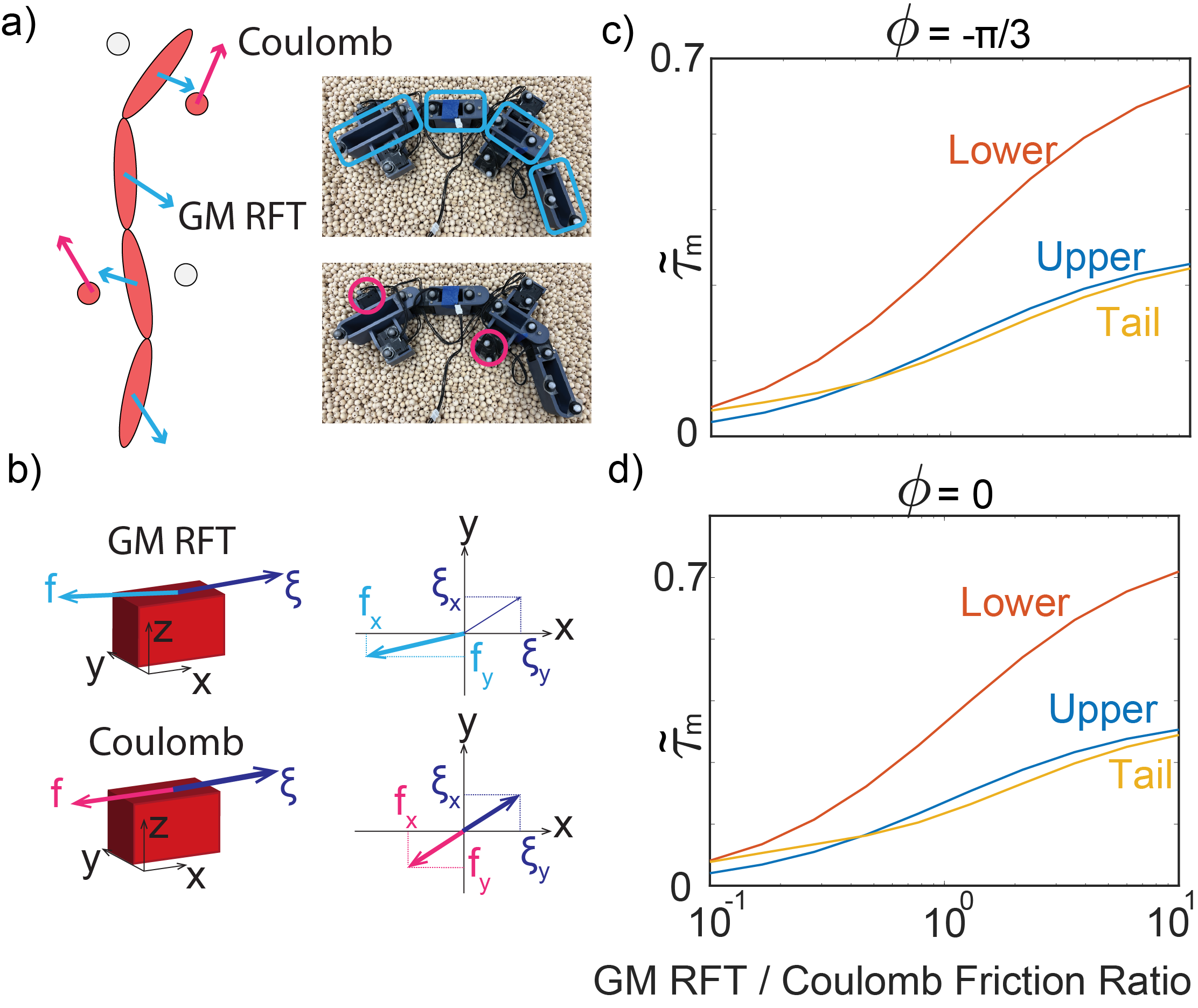}
    \vspace*{-5mm}
    \caption{\textbf{Model Prediction.} 
    (a) Left: Diagram of SILA Bot. Red shaded sections indicate contact with the ground. The magnitudes of Coulomb forces are indicated by magenta arrows, and Granular Media (GM) Resistive Force Theory (RFT) magnitudes are indicated by blue arrows. Right: Images of SILA Bot in GM. The bodies experiencing each type of friction at this time are highlighted in their respective colors. (b) An element experiencing GM RFT and another experiencing Coulomb Friction. The magnitudes of the forces are detailed in the plot. (c) Median torque, $\tilde{\tau}_m$, versus Ratio of GM RFT to Coulomb Friction across body joints at $\phi=-\pi/3$. (d) Median torque, $\tilde{\tau}_m$ versus Ratio of GM RFT to Coulomb Friction across body joints at $\phi=0$}
\label{fig:Model}
\end{figure}

Specifically, we assume that the robot's belly experiences reaction forces governed by GM RFT, while the feet interact with the substrate through dry Coulomb friction. In Fig.~\ref{fig:Model}, the reaction forces acting on the belly (blue vectors) and the feet (magenta vectors) are illustrated. Under Coulomb friction, the reaction force aligns opposite to the local slip velocity. In contrast, under GM RFT, the net reaction force depends on the orientation of the body segment relative to its direction of motion. In GM, drag is anisotropic: motion in the fore–aft direction encounters less resistance than lateral motion ($f_x < f_y$). This drag anisotropy is a key physical mechanism enabling limbless and undulatory robots to generate net thrust in granular substrates \cite{maladen2010SandSwimmingRobot, chong2023selfpropulsion, rieser2024GeometricPhase}.

Using the ground reaction force model and assuming quasi-static motion (i.e., instantaneous force and torque balance), we compute the body joint torques, $\tilde{\tau}$, as functions of the relative contributions of granular RFT and Coulomb friction. Torques are nondimensionalized by $\mu mg \cdot BL$, where $\mu$ is the friction coefficient, $m$ is body mass, $g$ is gravitational acceleration, and $BL$ is body length. As the relative contribution of granular RFT increases (corresponding to locomotion over deeper GM), the lower body joint experiences substantially higher torque compared to the upper and tail joints, approaching nearly twice their magnitude. 

We acknowledge that this model oversimplifies the dynamics of small-scale locomotion on shallow GM. Nevertheless, it captures a robust qualitative trend: during locomotion on deep granular substrates, joint torques become concentrated in the middle body segments, whereas on rigid flat ground, torques are distributed more uniformly along the body. This shift in torque distribution, rather than the precise force magnitude, provides informative structure for terrain differentiation and motivates the use of joint torque patterns as proprioceptive signals for substrate characterization.

\subsection{Terrain Characterization}

\begin{figure}
    \centering
    \includegraphics[width=\linewidth]{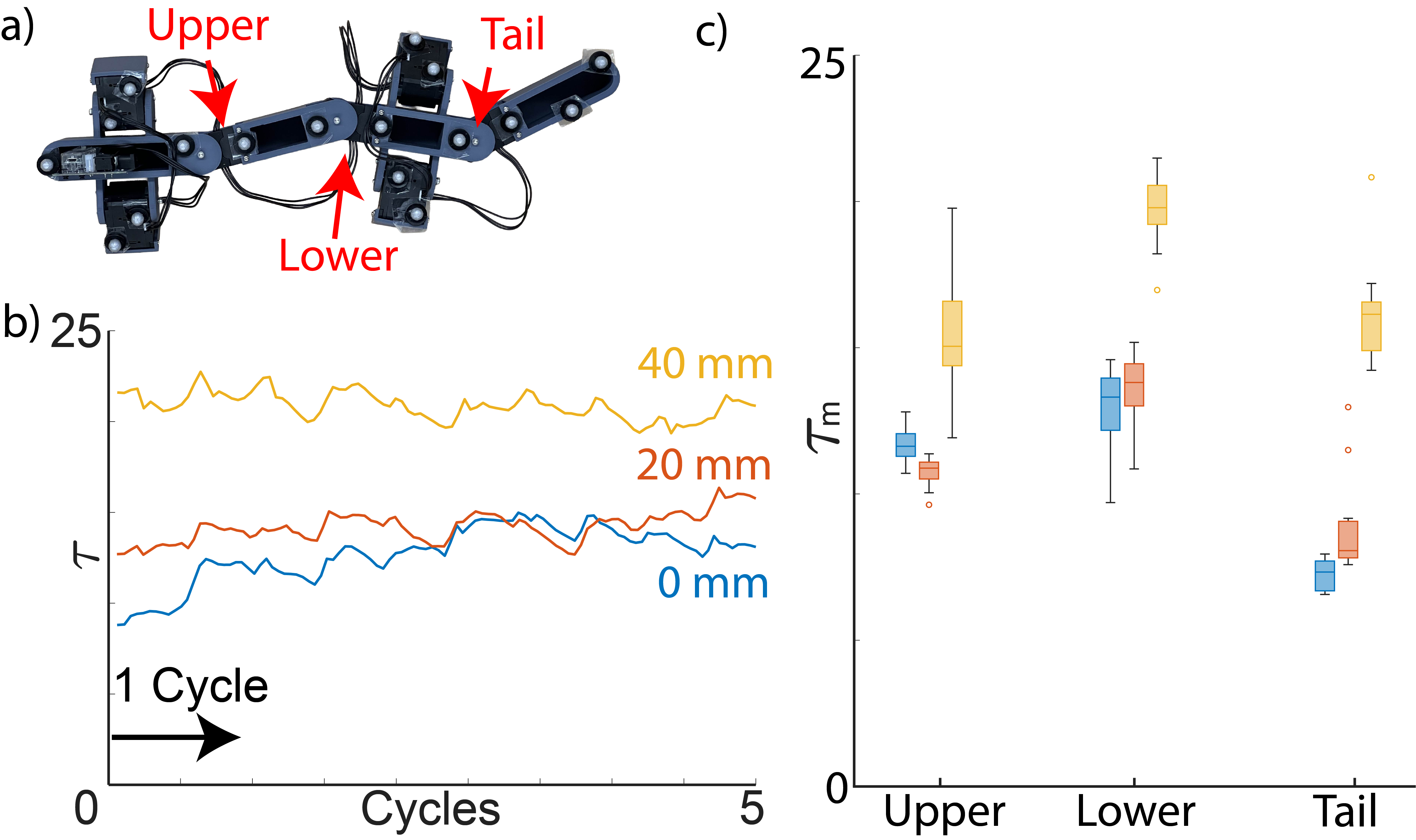}
    \vspace*{-5mm}
    \caption{\textbf{Proprioceptive sensing: Load estimation across granular media depths.} 
    (a) Three body motors on the SILA Bot (Upper, Lower, and Tail) provide estimates used for terrain characterization. 
    (b) Experimentally recorded load, $\tau$ of the lower body motor over five gait cycles for three granular media depths. Blue, orange, and yellow correspond to 0\,mm, 20\,mm, and 40\,mm depths, respectively. 
    (c) Summary of load measurements using the median value, $\tau_m$. The median $\tau$ is shown as a function of motor position (x-axis), with granular media depth color-coded as in (b).}
\label{fig:Load}
\end{figure}

Using the identification of optimal body phase angles, $\phi^{*}$, for 0, 20, and 40\,mm, depths, and inspired by model predictions, we seek to characterize the load data for the specified substrates. Depicted in Fig.~\ref{fig:Load}.a, we analyze Upper, Lower, and Tail motor load. We take the absolute value of the load to remove direction dependence (i.e., positive load is counter-clockwise, negative is clockwise): 
\begin{equation}
    \tau = \left| \tau_{raw} \right|
\end{equation}
where $\tau_{raw}$ is the raw data coming from the Dynamixel servo motors. The lower motor load is plotted as a timeseries in Fig. \ref{fig:Load}.b for SILA Bot's movement on 0, 20, and 40\,mm bead depth for 5 cycles. We notice that the motor load contains substantial variation within a cycle because of frequent direction changes. To minimize the effect of that variation, we take the median of $\tau$ across one cycle, defined as $\tau_m$. Using a box and whisker plot in Fig.~\ref{fig:Load}.c, we can compare the difference in terrain readings $\tau_m$ across motor positions and bead depth.

\begin{figure}
    \centering
    \includegraphics[width=\linewidth]{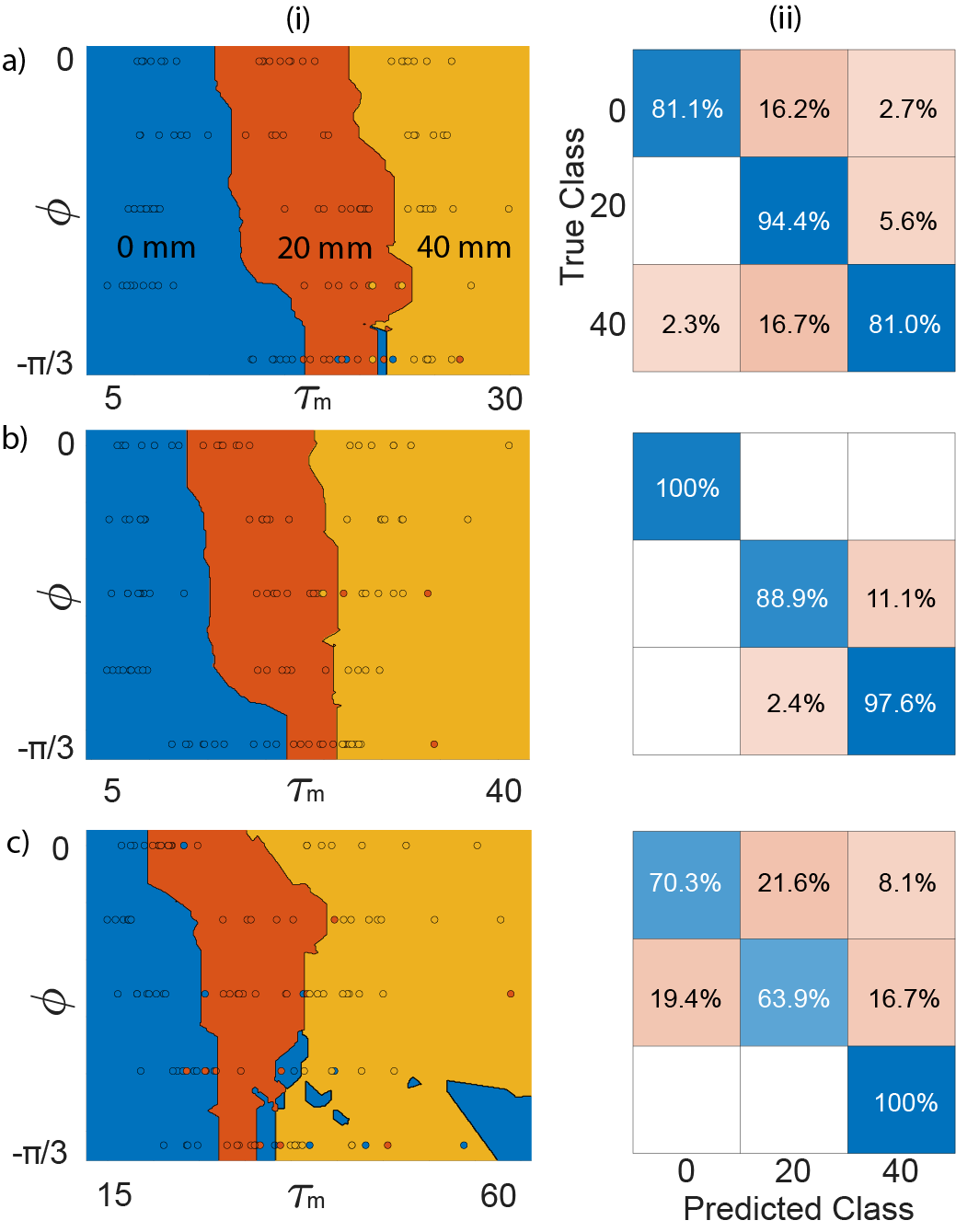}
    \vspace*{-5mm}
    \caption{\textbf{Terrain characterization via proprioceptive sensing.} 
    (i) Prediction of granular media depth using the robot gait parameter $\phi$ and the experimentally measured median load $\tau_m$. Terrain classification is performed using a $K$-Nearest Neighbors (KNN) model ($K=6$). The decision regions in the $\tau_m$-$\phi$ space are shown, where blue, orange, and yellow correspond to 0\,mm, 20\,mm, and 40\,mm depths, respectively. 
    (ii) Confusion matrices for terrain classification. For evaluation, 50\% of sampled $(\tau_m,\phi)$ pairs were used for training and the remaining 50\% for testing. 
    (a) Classification using upper body motor load estimates. The KNN decision map is fragmented (non-continuous regions), with an error rate up to $\sim$1/6. 
    (b) Classification using lower body motor load estimates. The KNN decision map is well-structured (continuous regions), with a low error rate of approximately 1/9. 
    (c) Classification using tail motor load estimates. The decision map is fragmented, with an error rate of approximately 1/5.}    
\label{fig:KNN}
\end{figure}

Notably, the value of $\tau_m$ also varies slightly with the commanded body phase offset, $\phi$. Combined, we now seek to estimate bead depth based on the commanded $\phi$ and the recorded $\tau_m$. We create a K-Nearest Neighbors (KNN) model (Euclidean distance, $k=6$) (Fig.~\ref{fig:KNN}) \cite{cover1967KNN}. In column (i), the decision map has median load $\tau_m$ on the $x$ axis and $\phi$ on the $y$ axis. We evaluate the accuracy of KNN by the confusion matrix in Fig.~\ref{fig:KNN} column (ii). We compare the KNN decision map and the model accuracy using (a) upper body, (b) lower body, and (c) tail motor load readings.

We observe that the decision regions obtained from the upper-body (Fig.~\ref{fig:KNN}.a) and tail (Fig.~\ref{fig:KNN}.c) load readings are not simply connected, which likely contributes to their lower classification accuracy. In contrast, the KNN decision map constructed from the mid-body joint load (Fig.~\ref{fig:KNN}.b) is simply connected and achieves an accuracy of 95\%. This structural separability motivates the use of the mid-body joint load for bead depth estimation. Because the $y$ axis, $\phi$, ultimately has a low impact on the prediction, we can find the transition points on the $x$ axis and approximate the bead depth as a linear function of median load.
We acknowledge that this depth estimate is coarse. However, given sensing noise and hardware limitations at small scales, we posit that even a rough estimate
is sufficient to construct a well-behaved feedback controller for terrain adaptation.

\subsection{Controller Design}

We have established that (1) within the experimentally relevant bead-depth range, the optimal body phase offset scales approximately linearly with bead depth, and (2) within this range, bead depth can be linearly estimated from motor load measurements. The composition of these two linear relationships naturally motivates a linear feedback controller that maps proprioceptive load signals directly to phase modulation. The linear feedback controller updates the body phase offset, $\phi$, at the end of each gait cycle (toward its optimal value):
\begin{equation}
    \phi(n+1)=\phi(n)+b_1(\tau_m(n)-\tau_0)-k(\phi(n)-\phi_0)
\end{equation}
where $\phi(n+1)$ is the updated body phase offset, $\phi(n)$ is the current phase, $\tau_m(n)$ is the median of the filtered load reading, and $b_1$, $k$, $\tau_0$, and $\phi_0$ are constants.

The constants $b_1$ and $k$ are experimentally determined ($b_1$ = -0.004 and $k$ = 0.005 for this paper). We set $\phi_0=-\pi/6$ because it is half the range of our desired body phase offsets, and serves to ``pull'' the value of $\phi(n+1)$ back to the middle, preventing it from growing too large in either direction. In a similar vein, we choose $\tau_0$ to be the mean of the load experienced when suspended in the air (no resistance from the substrate) and when in the highest complexity terrain (maximal resistance from the substrate). This provides an estimate of the middle level of load, such that flat, less complex, terrain loads will be lower than $\tau_0$ and deep GM loads will be higher. 


\subsection{Controller Validation in Consistent Terrain}
To validate the proposed control system, we conduct experiments with the SILA Bot on flat ground and in 40\,mm-deep wooden beads. Specifically, we aim to demonstrate that (1) the linear controller drives the body phase offset $\phi$ toward its terrain-dependent optimum $\phi^{*}$, and (2) the closed-loop controller ensures convergence of $\phi$ to $\phi^{*}$ under steady substrate conditions.

\begin{figure*}[t]
    \centering
    \includegraphics[width=\linewidth]{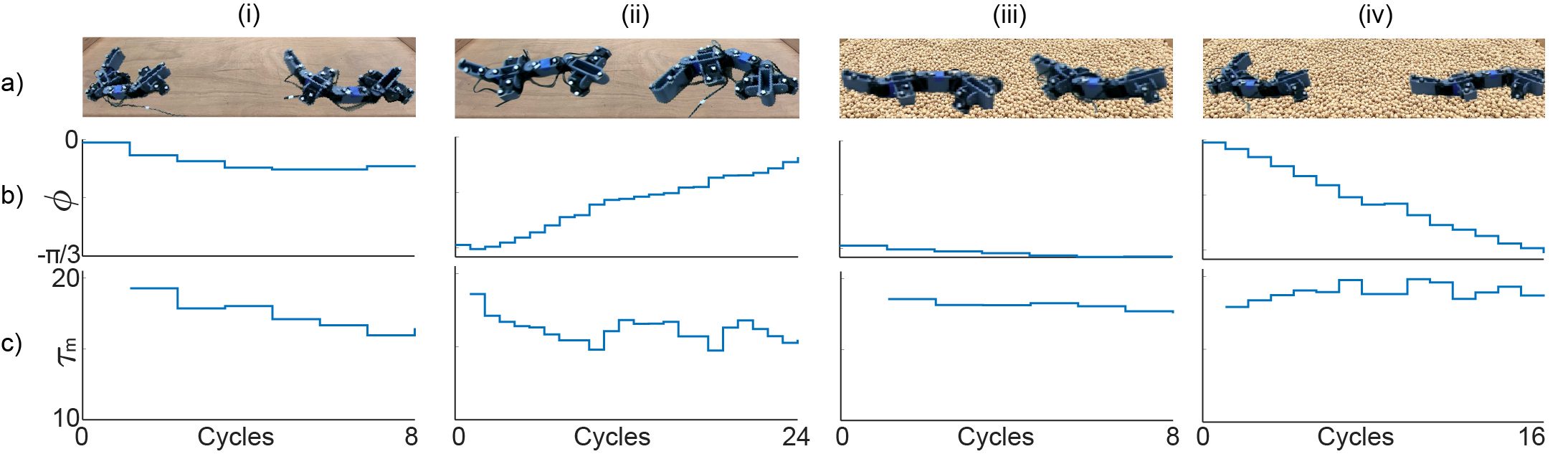}
    \vspace*{-6mm}
    \caption{\textbf{Using feedback control to adapt to desired body phase offset on consistent terrain.} The robot was placed onto both flat ground and granular media and tested. Row a) depicts the lizard robot at the beginning and end configurations across cycles. Row b) corresponds to the commanded body phase offset $\phi$. Row c) shows the median load as read by the robot that was used in the feedback controller to update $\phi$.
    Column (i) corresponds to a flat ground trial starting at $\phi=0$ and ending at $\phi\approx0$. (ii) starts on flat ground at $\phi=-pi/3$ and ends at $\phi\approx0$. (iii) starts in 40\,mm at $\phi=-pi/3$ and ends at $\phi\approx-pi/3$. (iv) starts in 40\,mm at $\phi=0$ and ends at $\phi\approx-pi/3$. }
\label{fig:ConstTerrain}
\end{figure*}

Columns Fig.~\ref{fig:ConstTerrain} (i) and (ii) represent 0\,mm while (iii) and (iv) correspond to 40\,mm deep wooden beads. To account for the limited size of the testing area, we command SILA Bot to walk 8 cycles. If the body phase offset has not stopped changing, we start another trial with the final body phase offset achieved from the previous trial to simulate as if it were walking uninterrupted in the environment. For example, column (ii) takes place over 3 trials of 8 cycles, which we concatenate into 24 cycles.

We conduct tests starting at both $\phi=0$ and $\phi=-\pi/3$ for each depth. Starting with flat ground, in b) (i), the body phase offset starts at $\phi=0$ and ends at $\phi\approx13\pi/180$\,rad. Then, in b) (ii) starting at $\phi=-\pi/3$, the final body phase offset is $\phi\approx10\pi/180$\,rad. Both of these angles are sufficiently close (within $\pi/12$\,rad) to the optimal body phase offset, $\phi^{*}=0$. Next, we test on 40\,mm deep. Starting at both $\phi=-\pi/3$ and $\phi=0$, in b) (iii) and (iv) respectively, SILA Bot achieves a final phase of sufficiently close to $\phi\approx-\pi/3$ (within $\pi/45$\,rad). 


\subsection{Controller Validation in Changing Terrain}
\begin{figure}
    \centering
    \includegraphics[width=\linewidth]{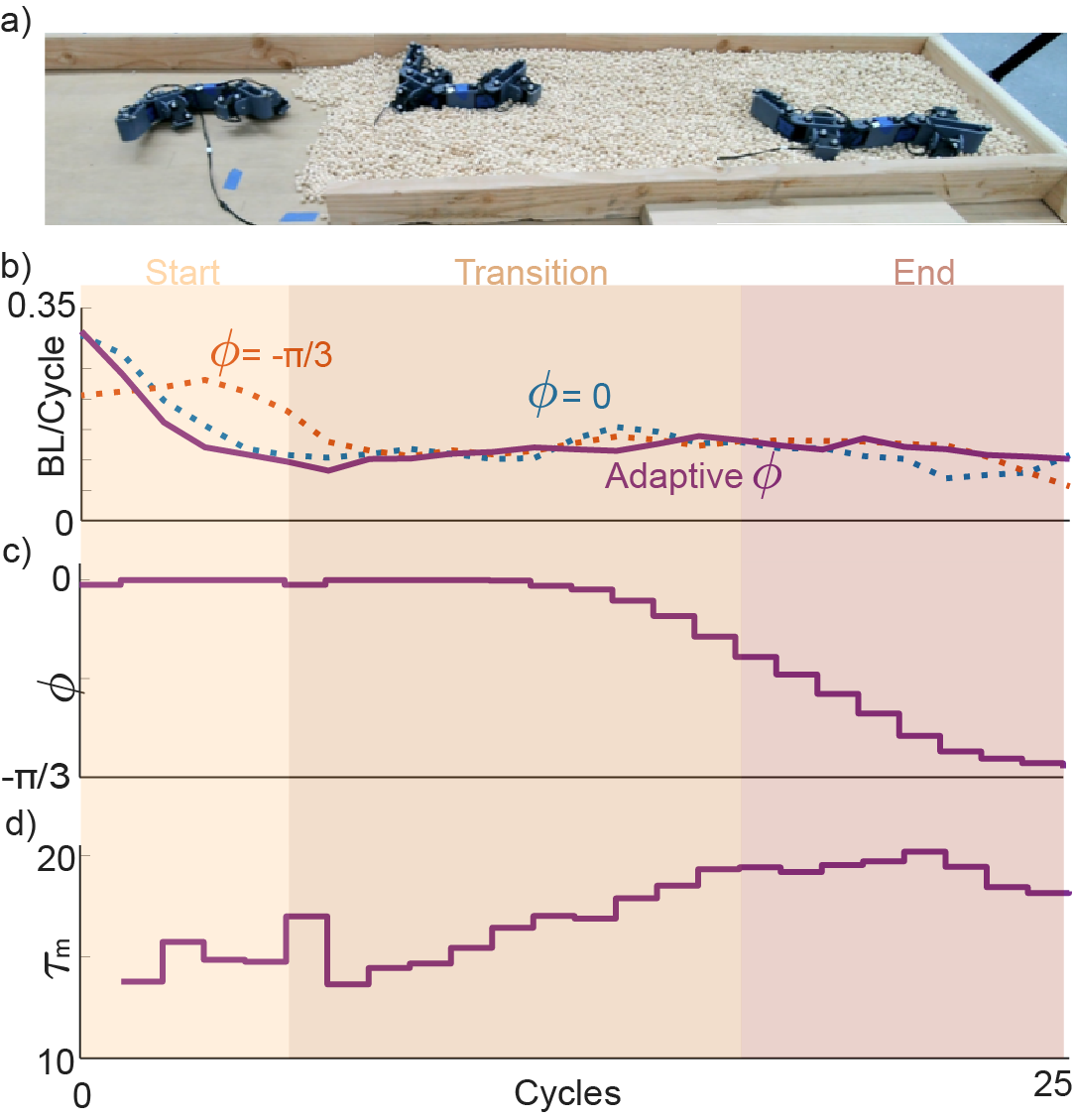}
    \vspace*{-5mm}
   \caption{\textbf{SILA Bot using feedback control to adapt body phase offset during terrain transitions.} 
    (a) Snapshots of the SILA Bot traversing terrain with unknown depth(transitioning from flat ground to sloped granular media and subsequently to deep granular media). 
    (b) Forward speed (units: body lengths per cycle, BL/C) as a function of time for three trials: adaptive $\phi$ control (solid purple), fixed $\phi = 0$ (dashed blue), and fixed $\phi = -\pi/3$ (dashed orange). The feedback controller achieves higher speed than $\phi = 0$ on flat terrain and higher speed than $\phi = -\pi/3$ on deep granular media, resulting in the highest overall average speed across terrain transitions.
    (c) Recorded body phase offset $\phi$ over time. The robot initializes at $\phi = 0$ and converges toward $\phi = -\pi/3$ as it enters deeper granular media. 
    (d) Median load $\tau_m$ recorded during locomotion and used as the feedback signal for phase adaptation. 
    Panels (b)–(d) are segmented into Start, Transition, and End sections corresponding to terrain changes.}
    \label{fig:Transition}
\end{figure}

To further evaluate the effectiveness of  of the feedback control, we seek to assess SILA Bot's performance as the terrain changes. For this test, we start on flat ground with a body phase offset of $\phi=0$ and aim the robot toward a section of wooden beads. The beads gradually slope up to approximately 40\,mm deep over a horizontal distance of $\approx60$\,cm. The test area setup is depicted in Fig.~\ref{fig:Transition}.a, which features snapshots of the robot at the Start on flat ground, the Transition period, and the End.

As shown in Fig.~\ref{fig:Transition}.b, the adaptive phase control system outperforms the feedforward control with fixed phase offsets $\phi = -\pi/3$ and $\phi = 0$ and successfully changes its body phase offset from $\phi=0$ to $\phi\approx-\pi/3$ over 25 cycles (Fig.~\ref{fig:Transition}.c). The transition occurs because of the increasing median load, $\tau_m$, as demonstrated in Fig.~\ref{fig:Transition}.d. Throughout the trial, SILA Bot was able to successfully change its body phase offset while locomoting over a changing substrate.
 
Under feedforward control with $\phi = 0$, the robot initially performs well on flat ground; however, its speed decreases substantially upon entering deep granular beads, eventually moving approximately 30\% slower than the adaptive case during cycles 22–25. Conversely, with feedforward control at $\phi = -\pi/3$, the robot performs well on deep beads but experiences a significant speed reduction when traversing flat ground, approximately 30\% slower than the adaptive case during cycles 1–3.

We suspect there may be speed losses that arise during the gait transition in the feedback controller. Large changes in body commanded phase offset can cause SILA Bot to pause in place for a moment. 

\section{CONCLUSION AND FUTURE WORK}

In this paper, we demonstrate that the optimal gait of the lizard-inspired robot, SILA Bot, is terrain-dependent. Specifically, when parameterized by the body phase offset $\phi$, the optimal value $\phi^*$ can be approximated as a linear function of granular media depth. We further measure the median load values, $\tau_m$, from the body servo motors and show that the lower body motor provides the most reliable indicator of granular depth, outperforming other body motors and achieving classification accuracy up to 95\%. Leveraging this proprioceptive signal and the depth–phase relationship, we design a simple linear feedback controller that adjusts the body phase offset toward its terrain-specific optimum. Experimental validation demonstrates that the controller autonomously estimates terrain depth and updates $\phi$ until converging to the optimal value $\phi^*$.

Our work provides a principled understanding of locomotion at the small scale (approximately 5\,cm), a regime that is fundamentally distinct from both macro- and micro-scale robotics. In large (e.g., $>50$\,cm) robots, high-bandwidth sensing and computation are energetically affordable and commonly used for perception-driven control. At the microscale (e.g., millimeter-scale robots), however, sensing can be energetically expensive and technically complex, often forcing systems to rely primarily on passive dynamics or minimal feedback. Small-scale robots occupy an intermediate regime: sensing is feasible but constrained in resolution, accuracy, and energy budget. This scale, therefore, demands a perception-control framework tailored to limited sensing resources. Our results demonstrate the effectiveness of proprioceptive sensing for terrain characterization and establish how such signals can be systematically mapped to adaptive locomotor strategies. 

Looking forward, this framework opens opportunities to co-design sensing systems, control policies, and robot morphology, enabling agile and energetically efficient small-scale robots capable of operating robustly in complex natural environments. For perception, future work will investigate: (1) how sensing accuracy and resolution can be improved under strict energy and hardware constraints; (2) how the proposed framework generalizes to other complex substrates, such as leaf litter, agricultural soils, and rubble. For control, future work will focus on: (1) faster convergence to the optimal gait under dynamic terrain transitions; and (2) developing a systematic analysis of robustness under environmental uncertainty, sensor noise, and actuator noise.

Finally, we are interested in understanding how sensing mechanisms should adapt when robot morphology changes, for example, through variations in limb length, limb posture, or overall body geometry. It is possible that different morphological designs may favor different distributions of sensing and actuation. For instance, a robot with multiple redundant motors but minimal sensing may achieve performance comparable to that of a robot with fewer, higher-precision actuators supported by richer sensory feedback. Ultimately, this perspective motivates a systematic comparison between the cost of computation and the cost of actuation, enabling quantitative evaluation of trade-offs toward achieving robust and energetically efficient locomotion.

\section*{ACKNOWLEDGMENTS}
The authors would like to acknowledge Jackson Habala and Xiyuan Wang for their feedback on the paper.


\addtolength{\textheight}{-12cm}   





\bibliographystyle{ieeetr}
\bibliography{bib}

\end{document}